\documentclass[lettersize,journal]{IEEEtran}
\usepackage{amsmath,amsfonts}
\usepackage{algorithmic}
\usepackage{array}
\usepackage[caption=false,font=normalsize,labelfont=sf,textfont=sf]{subfig}
\usepackage{textcomp}
\usepackage{stfloats}
\usepackage{url}
\usepackage{verbatim}
\usepackage{graphicx}
 \usepackage{siunitx} 
\def\BibTeX{{\rm B\kern-.05em{\sc i\kern-.025em b}\kern-.08em
    T\kern-.1667em\lower.7ex\hbox{E}\kern-.125emX}}
\usepackage{balance}
\begin{document}

\title{
Effect of Twisted-Yarn Architecture on Pressure and Proximity Sensing Characteristics of Textile Capacitive Sensors for Robotic Skin}
\author{Ishtia Zahir,
Eslam Saleh,
Maryam Rezayati,
Günter Grabher,
and Gaffar Hossain
\thanks{Ishtia Zahir, Eslam Saleh, and Gaffar Hossain are with V-Trion GmbH, Millennium Park 15, 6890 Lustenau, Austria. Maryam Rezayati is with the Institute of Mechatronic Systems, ZHAW Zurich University of Applied Sciences, Winterthur, Switzerland. Günter Grabher is with Grabher Group GmbH, Augarten Strasse 27, 6890 Lustenau, Austria.}
\thanks{Corresponding authors: Gaffar Hossain (g.hossain@v-trion.at) and Maryam Rezayati (rzma@zhaw.ch).}
}
\markboth{ Submitted to IEEE Transactions On Instrumentation and Measurement}%
{Submitted to  IEEE Transactions On Instrumentation and Measurement}

\maketitle

\begin{abstract}
Textile-integrated capacitive sensors offer flexible and conformable tactile sensing for wearable electronics and human--robot interaction; however, the influence of yarn-level architecture on capacitive transduction characteristics remains insufficiently quantified. 
This work presents a textile capacitive sensing platform based on silver-coated yarns coated with polydimethylsiloxane and assembled into one-, two-, and four-layer twisted configurations. 
The influence of effective electrode overlap area and inter-fiber separation on the capacitive response is systematically investigated, enabling architecture-dependent tuning of pressure and proximity sensing characteristics. Pressure was calculated using the localized single-fiber contact area, corresponding to stresses of 0.4–3.9~MPa.
Increasing the layer number improved mechanical strength and sensing performance: elongation at break increased from 37.5\% to 62.5\% and 85.0\%, while the maximum load increased from 23.3 to 42.7 and 89.7~N. Sensitivity increased with layer number and frequency, reaching 0.1331~MPa$^{-1}$ for the four-layer sensor at 100~kHz. The four-layer configuration also exhibited low hysteresis, minimal thermal drift from 25 to 90~$^\circ${C}, and stable operation over 15,000 cycles. Proximity detection ranges of 60, 50, and 
40~mm were obtained for the one-, two-, and four-layer sensors, respectively, revealing an architecture-dependent sensitivity-range trade-off. 
A 4$\times$4 textile sensing array enabled spatial contact mapping, while robotic-arm integration demonstrated real-time touch and proximity detection with an end-to-end robotic system latency (from detection to robot reaction) of 403~ms. The results establish yarn architecture as a tunable design parameter governing the measurement characteristics of textile-integrated capacitive sensing systems.
\end{abstract}


\section{Introduction}
Reliable measurement of contact pressure and object proximity on flexible and curved surfaces is important for wearable electronics, electronic skin, soft robotics, and human–robot interaction systems\cite{Cho2022}. Among the various transduction mechanisms, including piezoresistive, piezoelectric, triboelectric, and capacitive, capacitive pressure sensors have attracted particular attention due to their low power consumption,  stable baseline  response, and ability to support both contact and pre-contact detection through changes in the electric field \cite{Ha2022, Niu2024}. 
Electronic textiles provide a promising platform for such systems because they are flexible, breathable, lightweight, and conformable to complex three-dimensional surfaces \cite{Cho2022}. Recent studies have demonstrated textile-based capacitive and piezoresistive sensors for plantar-pressure measurement, gait-phase detection, and lower-limb motion monitoring \cite{Tan2021, Wu2025}. Textile capacitive sensors generally employ sandwich-type, in-plane, or yarn-based architectures \cite{su2022}. Sandwich structures are widely used, but their  response depends strongly on the compressibility and mechanical stability of the dielectric layer. In-plane structures, including interdigitated electrodes, offer surface-integrated sensing but can involve complex patterning and limited scalability over large textile areas. Yarn-based architectures instead form localized capacitive junctions between dielectric-coated conductive fibers, making the fiber–fiber junction, rather than the macroscopic textile area, the active sensing element. 
Recent work has demonstrated highly sensitive capacitive fiber pressure sensors through regulation of the electrode configuration and dielectric layer \cite{Qu2023}. Helical auxetic yarns have also been investigated as capacitive strain sensors, demonstrating that yarn geometry can strongly influence electromechanical sensitivity \cite{Cuthbert2023}. More recent flat-knitted and cut-pile textile capacitive sensors have further shown the potential for scalable fabrication and direct integration into wearable structures \cite{Fischer2024, Motaghedi2025}. However, these studies have primarily focused on material optimization, fabric-level architectures, or strain sensing. The effects of twisted-yarn layer number and multilayer helical arrangement on pressure and proximity sensing at discrete fiber–fiber junctions therefore remain insufficiently quantified.
Understanding the role of these structural parameters is essential because the arrangement of conductive fibers directly controls the effective electrode overlap area and the inter-fiber separation distance at the contact junction, which together determine the capacitance response.  A systematic structure–measurement relationship is therefore required to explain how yarn-layer architecture affects pressure sensitivity, proximity range, hysteresis, response time, thermal stability, repeatability, and long-term durability. Such a framework would enable sensor performance to be tuned through geometry without changing the constituent materials or introducing additional fabrication steps.
Dual-mode sensing is particularly important in robotic e-skin because proximity information can support pre-contact awareness, while tactile information provides feedback after physical interaction \cite{Wu2025, su2022}.  Flexible multimodal networks and hybrid pressure–proximity sensors have been demonstrated on robotic platforms \cite{Ham2022, Ge2021}, but their fabrication and integration approaches remain largely device- and surface-specific.  Textile platforms offer an attractive alternative because they can conform to large, curved, and deformable surfaces while preserving flexibility and scalability.
Our group has previously demonstrated a fiber-based piezoresistive sensing platform using graphite–polyurethane-coated conductive yarns in different ply configurations, showing that layer architecture governs sensitivity, hysteresis, and durability \cite{Zahir2026}. However, piezoresistive systems can be susceptible to  temperature-dependent resistance variation and do not inherently provide proximity sensing. These limitations motivate the development of a capacitive counterpart that preserves the same scalable fiber-based geometry while enabling dual-mode contact and pre-contact sensing.
Here, we present a textile-integrated capacitive sensing platform that systematically investigates twisted-yarn layer architecture as the principal structural design parameter. Polydimethylsiloxane-coated silver yarns are assembled into 1-layer, 2-layer, and 4-layer twisted configurations and integrated into a grid-based textile structure.  We show that the capacitive response is governed by the interplay between effective electrode overlap area and inter-fiber separation at the fiber–fiber contact junction, both of which are modulated by the multilayer helical yarn geometry. Pressure is referenced to the localized single-fiber contact area rather than the macroscopic textile area, thereby representing the mechanical stress acting directly at the active sensing region. 
The proposed architecture enables pressure sensitivity and proximity range to be tuned without changing the material composition. Increasing the layer number improved both mechanical performance and pressure sensitivity, with the 4-layer sensor reaching a maximum sensitivity of $0.1331~\mathrm{MPa}^{-1}$ at 100 kHz. The developed platform exhibits stable cyclic performance over 15,000 loading cycles, low hysteresis, and minimal temperature-induced drift from 25 to 90~$^\circ\mathrm{C}$. In addition, the system enables proximity detection of approaching objects through fringe electric-field perturbation, with detection ranges of 60, 50, and 40 mm for the 1-layer, 2-layer, and 4-layer configurations, respectively. These results reveal an architecture-dependent sensitivity–range trade-off: increasing the layer number improves contact-pressure sensitivity through increased effective electrode overlap and structural deformation, while simultaneously confining the fringe electric field and reducing the pre-contact detection range. A $4\times4$ textile sensing matrix demonstrates spatial mapping of single- and multi-point contact events. Furthermore, different sensor implementations were integrated onto a robotic platform to demonstrate real-time proximity and touch detection for human–robot interaction, achieving an end-to-end robotic system latency of 403 ms.

The core contributions of this paper are summarized as follows:
\begin{itemize}
    \item A systematic investigation of how 1-layer, 2-layer, and 4-layer twisted-yarn architectures influence effective electrode overlap, inter-fiber separation, pressure sensitivity, and proximity range. 
    \item Establishment of an architecture-dependent structure–measurement relationship that enables sensing characteristics to be tuned through yarn geometry without modifying the constituent materials.
    
    \item Comprehensive characterization of pressure response, mechanical properties, hysteresis, dynamic behavior, thermal stability, repeatability, and durability over 15,000 loading cycles, with pressure referenced to the localized active fiber-junction area.
    
    \item Identification of an architecture-dependent trade-off between contact-pressure sensitivity and pre-contact detection range, followed by system-level validation using a 4$\times$4 textile sensing matrix.

    \item System-level deployment through three robotic-arm demonstrations comprising touch localization, proximity-triggered collision avoidance, and object-aware safety monitoring.
    
\end{itemize}
\section{Related Work}
Textile capacitive pressure sensors commonly use sandwich-type electrode–dielectric–electrode structures because of their simple design and compatibility with large-area fabrication \cite{su2022}. However, their performance depends strongly on the compressibility and mechanical stability of the dielectric layer. Fiber- and yarn-based configurations provide an alternative by localizing the sensing response at discrete fiber–fiber or yarn–yarn interfaces.

Recent studies have demonstrated that fiber architecture significantly influences sensing performance. Sun et al. developed a machine-braided iontronic pressure-sensing yarn with high sensitivity and a broad pressure-detection range \cite{Sun2026}. Qi et al. reported a helical liquid-metal nanofiber yarn for resistive strain sensing, demonstrating that helical organization can improve stretchability and sensing stability \cite{Qi2025}. Liu et al. further showed that fiber diameter, porosity, and internal microstructure affect the performance of fiber-based iontronic pressure sensors \cite{Liu2025}. 
Although these studies demonstrate the importance of fiber-level structure, the influence of twisted-yarn layer architecture on effective electrode overlap area, inter-fiber separation, and the resulting capacitive sensing characteristics remains insufficiently quantified.

Combined pressure and proximity sensing is particularly important for robotic electronic skin, where proximity sensing provides pre-contact awareness and pressure sensing supplies tactile feedback after contact \cite{Niu2024, Wu2024, Ge2021}. Ye et al. demonstrated an all-fabric capacitive sensor capable of pressure and noncontact detection and implemented it as a $4\times4$ sensing array \cite{Ye2022}. Rashid et al. developed a multifunctional textile–plastic composite providing pressure mapping, proximity detection, and heating \cite{Rashid2025}. 
However, these systems primarily employ fabric-level architectures or separate sensing elements. A quantitative understanding of how multilayer twisted-yarn architectures influence effective electrode overlap area, inter-fiber separation, pressure sensitivity, and proximity response within a unified textile capacitive sensing platform remains limited. Furthermore, the validation of such architecture-engineered textile sensing systems in functional robotic applications has received comparatively less attention. Addressing these gaps is the primary objective of the present work.

\section{Material and Methods} \label{section_3}
\subsection{Sensor Design and Scalable Fabrication}
The proposed capacitive sensing platform employs a core–sheath fiber architecture integrated into a flexible textile grid \cite{su2022, Qu2023, Motaghedi2025, Ye2022}. Silver-coated yarns (Shieldex 117/17 dtex HC, Statex, Germany), with a nominal diameter of approximately 0.3 mm and a linear resistance of approximately \SI{30}{\ohm\per\centi\meter}, were selected as the conductive core. These yarns combine high electrical conductivity with sufficient mechanical robustness for processing using conventional textile machinery \cite{Cho2022, su2022, Fischer2024, Motaghedi2025}.
A scalable, continuous nozzle-based coating process was used to form the dielectric sheath. The conductive yarn was fed at a controlled speed of \SI{10}{\meter\per\minute} through a coating nozzle containing uncured polydimethylsiloxane (PDMS; Sylgard 184, Sigma-Aldrich). The coated yarn was then passed through a cylindrical heating chamber maintained at 80 ~$^\circ\mathrm{C}$ to thermally cure the polymer, producing a uniform dielectric layer approximately 0.2 mm thick. PDMS was selected because of its flexibility, water resistance, chemical inertness, low thermal expansion, and well-characterized elastic properties \cite{Ha2022, su2022, Qu2023, Ye2022}. The compliant coating protects the conductive yarn without substantially increasing fabric stiffness and enables rapid elastic recovery after compression. These characteristics support reversible deformation at the sensing junctions, stable dielectric behavior under mechanical loading, and reduced temperature-induced signal drift \cite{Ha2022, su2022, Qu2023}.
To systematically investigate the influence of yarn-level architecture, these dielectric-coated fibers were assembled into single-layer (1-layer), two-layer (2-layer), and four-layer (4-layer) configurations by twisting the corresponding number of fibers together. A key mechanical benefit of this twisted helical structure is that it locks the fibers tightly together. This prevents the individual threads from slipping against each other under pressure, reducing relative fiber displacement during loading and improving structural integrity during repeated operation \cite{Cuthbert2023, Sun2026, Qi2025}.
Finally, to construct the complete sensing matrix, these customized yarns were stitched onto a woven polyester substrate in a grid configuration using conductive thread. The row and column yarns were arranged orthogonally, with each intersection acting as an individual capacitive sensing node \cite{Milovic2022, Wu2025, Fischer2024}. The grid was designed with a pitch of approximately 10 mm, resulting in an active sensing area of 30 × 30 mm², and external electrical connections were securely established at the yarn terminals using silver conductive adhesive and copper wires.

\subsection{Measurement Principle and Stress Normalization}\label{sp} 
\begin{figure*}
\centerline{\includegraphics[width=\textwidth]{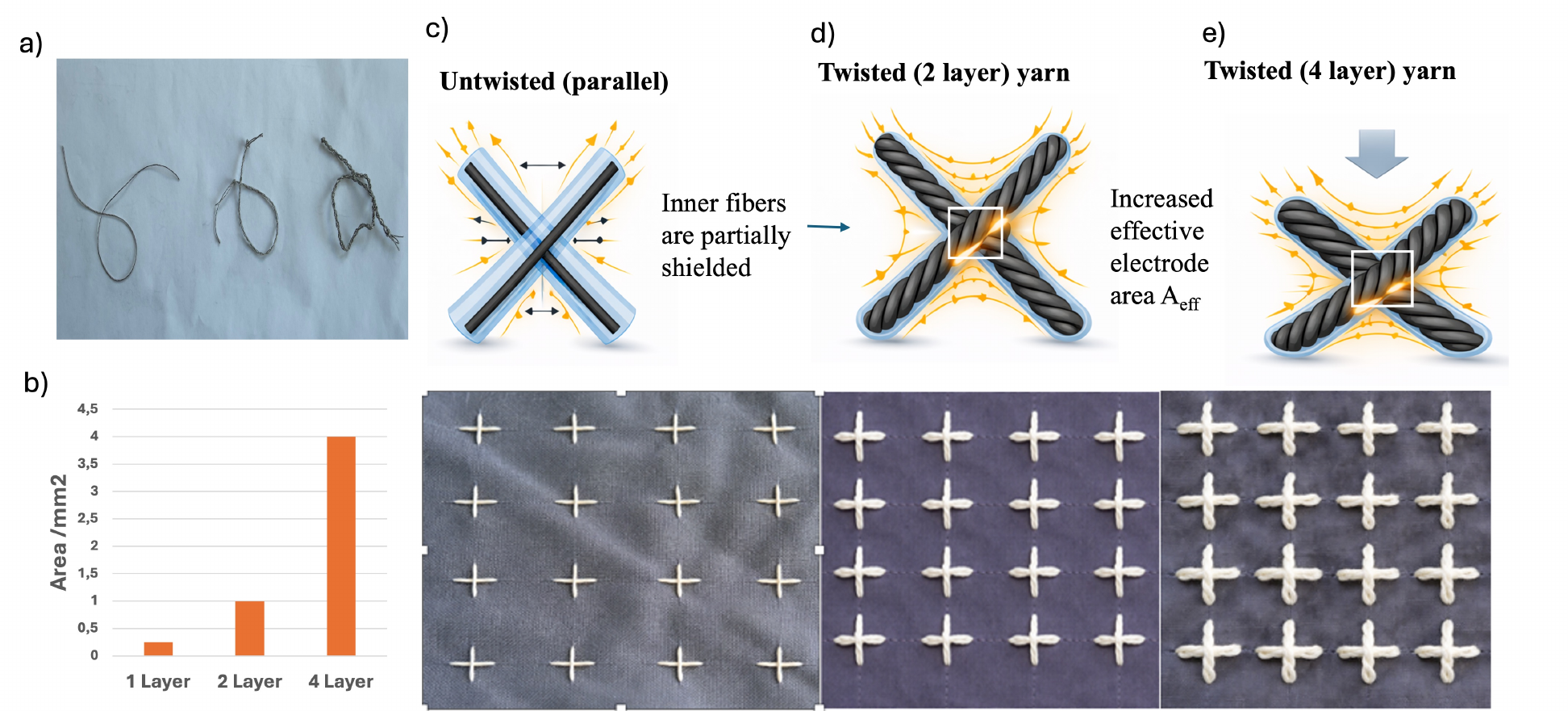}}
\caption{Textile capacitive sensor characterization. 
(a) One-, two-, and four-layer twisted yarns. 
(b) Corresponding $A_{\mathrm{eff}}$ values. 
(c--e) Junction schematics and textile grids showing field distribution, larger junction area, and reduced $d_{\mathrm{eff}}$.}
\label{fig:characterization}
\end{figure*}

\subsubsection{Sensing Mechanism}
The operating principle of the textile capacitive sensor is based on pressure-induced modulation of capacitance between conductive fiber electrodes separated by a deformable PDMS dielectric layer \cite{Ha2022, su2022, Qu2023}. The capacitance response can be approximated using the classical capacitance relationship:
\begin{equation}
C = \frac{\varepsilon_0 \varepsilon_r A_{\mathrm{eff}}}{d_{\mathrm{eff}}}
\label{eq:capacitance}
\end{equation}
where  $\varepsilon_0$  is the free space permittivity,  $\varepsilon_r$ is the relative permittivity of PDMS, $A_{\mathrm{eff}}$  is the effective electrode overlap area, and $d_{\mathrm{eff}}$ is the inter-fiber separation at the contact junction. Fig.~\ref{fig:characterization} shows the fabricated 1-layer, 2-layer, and 4-layer twisted yarn samples. PDMS provides elastic recovery, chemical stability, and stable dielectric properties, which support reversible capacitive sensing under repeated loading \cite{Ha2022, su2022, Qu2023}.
The yarn architecture strongly influences the sensor response. In the 1-layer structure, the fibers form limited contact junctions, resulting in a smaller $A_{\mathrm{eff}}$ and a wider fringe electric field. Twisting the yarns into 2-layer and 4-layer structures increases the contact/overlap area and reduces the effective fiber separation. Under applied pressure, the junction is compressed, $d_{\mathrm{eff}}$ decreases, and $A_{\mathrm{eff}}$ increases, 
leading to a higher capacitance. The measured effective contact area increases from 0.25 mm² for the 1-layer yarn to 1.0 mm² for the 2-layer yarn and 4.0 mm² for the 4-layer yarn, giving a 16-fold increase from 1-layer to 4-layer. However, the denser multilayer structure also confines the fringe field, reducing the proximity detection range from 60 mm to 50 mm and 40 mm for the 1-layer, 2-layer, and 4-layer samples, respectively. This indicates a trade-off between pressure sensitivity and proximity sensing range.

\subsection{Sensor Characterization}
To quantify the sensing and measurement characteristics of the proposed platform, 
the sensors were systematically characterized using a multi-domain approach. Specifically, this evaluation encompasses fundamental electrical stability via frequency-dependent analysis, environmental resilience via thermal stability monitoring, mechanical robustness through static and cyclic contact pressure testing, and spatial awareness through non-contact proximity detection.

\subsubsection{Frequency-Dependent Capacitive Response} 
To evaluate the influence of electrical excitation frequency on sensor performance, initial capacitance measurements were performed using a precision LCR meter (Keysight U1733C) with an AC excitation voltage of 1 V. Measurements were systematically conducted at 1 kHz, 10 kHz, and 100 kHz. The results showed that the magnitude of the normalized capacitance response varied with measurement frequency, confirming frequency-dependent sensitivity across the investigated range \cite{su2022, Qu2023}.

\subsubsection{Thermal Stability} 
Because traditional flexible sensors often suffer from high susceptibility to environmen-tal noise, the thermal stability of the sensor was characterized \cite{Ha2022, su2022}. Temperature stability was evaluated by placing the sensor in a temperature-controlled chamber and varying the ambient temperature from 25~$^\circ\mathrm{C}$  to 90~$^\circ\mathrm{C}$  while continuously monitoring the capacitance output. This confirmed the system’s resistance to temperature-induced drift prior to practical implementation.

\subsubsection{Contact Pressure Characterization} 
Pressure-dependent measurements were performed by applying calibrated masses to the sensor while continuously recording the capacitance response. The applied pressure was calculated using the localized single-fiber contact area of approximately 0.07 mm² rather than the macroscopic area of the textile grid. This approach represents the actual mechanical stress concentrated at the active fiber-to-fiber junction. The applied masses corresponded to contact pressures of 0.4, 0.8, 2.0, and 3.9$~\mathrm{MPa}$, estimated using the cylindrical-fiber contact geometry and Hertzian contact mechanics \cite{Bunge2021}. This normalization approach enables pressure characterization to be referenced to the active sensing junction, providing a physically meaningful basis for comparing architecture-dependent sensing characteristics.
Hysteresis was evaluated under two complementary loading conditions. Dynamic hysteresis was measured during continuous loading–unloading over a pressure range below sensor saturation. Quasi-static hysteresis was assessed by applying and removing discrete masses up to 1 kg. These tests enabled comparison of the loading and unloading responses and evaluation of mechanical recovery.
Repeatability was assessed under continuous cyclic loading using a 100 g mass, while long-term durability was evaluated over 15,000 loading–unloading cycles using a motorized linear stage equipped with a load cell. Response and recovery times were extracted from transient capacitance signals during loading and unloading. The dynamic response was further evaluated by applying periodic mechanical excitation at frequencies from 1 to 5 Hz.

\subsubsection{Mechanical and Sensitivity Characterization}
Mechanical characterization of the 1-, 2-, and 4-ply textile sensors was performed using a motorized tensile/compression testing system equipped with a load cell and a vertically controlled displacement stage, as shown in Fig.~\ref{figure_2}(i). For tensile testing, the sensor specimens were fixed between the grips and stretched at a controlled rate until failure to determine elongation at break and stretchability. For compression measurements, the sensor was positioned on the lower platform and subjected to controlled localized loading, while the corresponding force, displacement, and capacitance response were recorded. Pressure sensitivity was subsequently evaluated at 1, 10, and 100 kHz using an LCR meter with a 1~V~AC excitation \cite{Ha2022,su2022,Qu2023,Bunge2021}.

\subsubsection{Non-Contact Proximity Characterization} 
In addition to physical contact, the sensor’s dual-mode, precontact capabilities were evaluated \cite{Wu2024, Motaghedi2025, Ge2021, Ye2022}. Objects with varying dielectric properties (a human hand, metal plate, plastic sheet, and paper) were positioned at controlled distances of 0-60 mm from the sensor surface, and the capacitance response was recorded at 1 kHz. The maximum detectable distance was determined independently for each yarn configuration, yielding detection ranges of 60, 50, and 40 mm for the 1-layer, 2-layer, and 4-layer configurations, respectively. 
\subsection{System Integration and Robotic Validation}
A 4$\times$4 textile sensor matrix was interfaced with a PSoC 6 microcontroller using the integrated CapSense module. The row and column yarns were connected to independent sensing channels, and the acquired capacitance data were transmitted through USB–UART to a graphical user interface. The signals were visualized in real time as 2D heatmaps and 3D surface plots to indicate the location and intensity of touch or proximity events.
For the robotic demonstrations, a flexible yarn sensor was attached to a Franka Emika Panda robot without restricting its motion. Capacitance was measured using an FDC1004 converter and M5STACK microcontroller connected to a PC through serial communication. Localized touch inputs were mapped to directional robot commands for intuitive control, while predefined capacitance thresholds were utilized to trigger autonomous evasion maneuvers and classify objects for context-aware safety stops.

\section{Results and Discussion}

\subsection{Pressure Response and Mechanical Characterization}
\begin{figure*}
\centerline{\includegraphics[width=\textwidth]{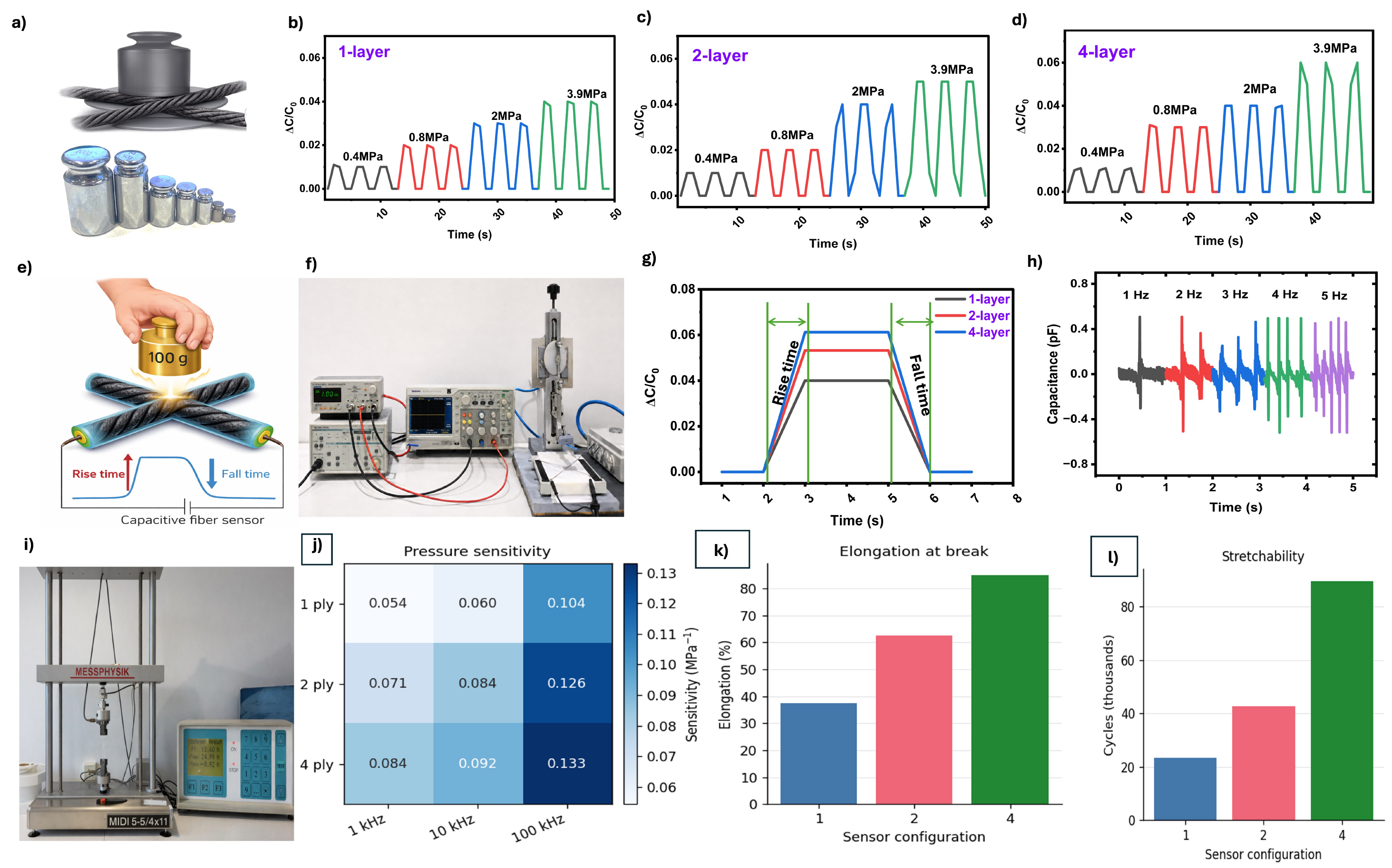}}
\caption{Pressure, frequency, and mechanical characterization of the textile capacitive sensor. 
(a--d) Stepwise pressure response. 
(e--h) Setup, response time, and 1--5 Hz frequency behavior. 
(i--l) Tensile properties and frequency-dependent sensitivity.}
\label{figure_2}
\end{figure*}

Fig.~\ref{figure_2}(a–d) shows the stepwise pressure-dependent capacitance response ($\Delta C / C_0$) of 
the 1-layer, 2-layer, and 4-layer configurations under static applied pressures of 0.4, 0.8, 2.0, and 3.9$~\mathrm{MPa}$, calculated at the fiber–fiber contact junction as described in Section \ref{section_3}.1. All results exhibit well-defined and stable capacitance plateaus at each pressure level, with $\Delta C / C_0$ increasing monotonically with applied pressure and returning consistently to baseline upon unloading, while the 4-layer produces the highest capacitance change at all pressure levels, confirming that the layer-dependent enhancement is maintained across the full pressure range.

A 100 g mass applied to the sensor surface yielded rise and recovery times of approximately 0.9 s across all layer configurations as shown in Fig.~\ref{figure_2}(g). 
The frequency dependent sensing behavior was characterized under periodic mechanical excitation at 1–5 Hz, as shown in Fig.~\ref{figure_2}(h). These measurements demonstrate detection of periodic excitation rather than full settling of the sensor response within each cycle. All configurations exhibit stable and well-resolved output signals across the investigated frequency range with no observable attenuation, baseline drift, or waveform distortion, confirming reliable tracking of dynamic stimuli within the frequency range relevant to human motion and wearable sensing applications \cite{Milovic2022, Wu2025}. 
The mechanical performance of each yarn configuration was evaluated using a tensile testing system, as shown in Fig.~\ref{figure_2}(i). The elongation at break increased from 37.5\% for the one-layer configuration to 62.5\% and 85.0\% for the two- and four-layer configurations, respectively. The maximum load increased correspondingly from 23.3 to 42.7 and 89.7 N. Sensitivity increased with both layer number and measurement frequency, reaching a maximum of 0.1331~$\mathrm{MPa}^{-1}$  for the four-layer sensor at 100 kHz.

\subsection{Repeatability and Durability}
\begin{figure*}
\centering
\includegraphics[width=\textwidth]{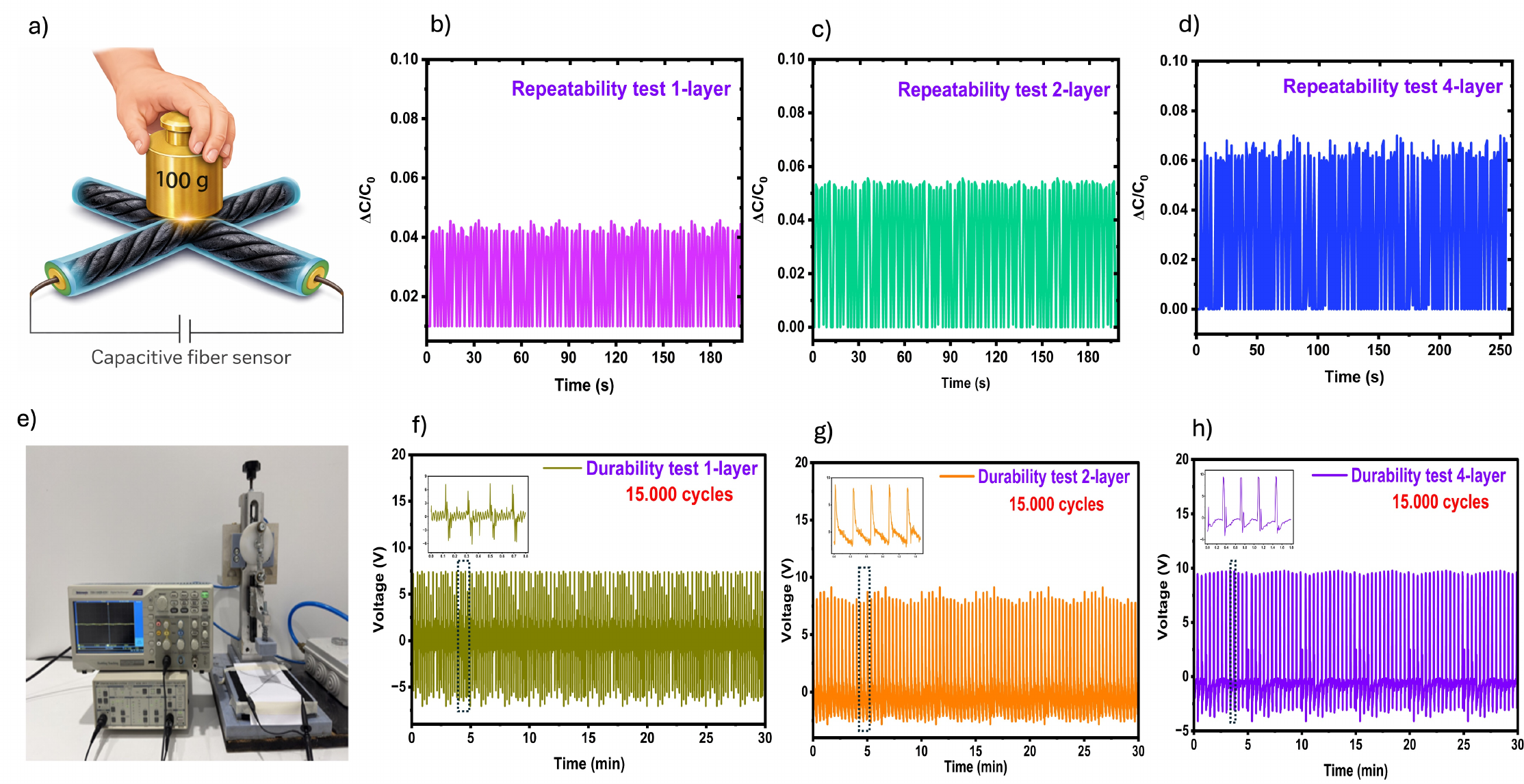}
\caption{Repeatability and durability of the textile capacitive sensor. 
(a) Cyclic loading schematic. 
(b--d) Repeatability over 200 s. 
(e) Durability test setup. 
(f--h) Long-term response over 15,000 cycles.}
\label{figure_3}
\end{figure*}

The repeatability of the sensor was evaluated under continuous cyclic loading with a 100g mass, as shown in Fig.~\ref{figure_3}(b–d). All configurations maintained stable and consistent $\Delta C / C_0$ amplitudes throughout the test, with negligible signal drift or baseline shift. The four-layer configuration exhibited the highest response amplitude, followed by the two- and one-layer configurations, confirming both layer-dependent sensitivity and reliable repeatability.
Long-term durability was assessed over 15,000 loading–unloading cycles using the setup in Fig.~\ref{figure_3}(e), with the corresponding results shown in Fig.~\ref{figure_3}(f–h). All configurations retained stable responses without observable degradation over the investigated 15,000-cycle test, confirming the durability of the sensing platform.

\subsection{Dynamic Pressure Response and Hysteresis}

Hysteresis was evaluated under two complementary loading conditions. In the dynamic test, the sensor was continuously loaded and unloaded over a pressure range below saturation. In the quasi-static test, discrete masses were applied and removed up to 1 kg, allowing the loading and unloading responses to be compared across the full operating range.

\begin{figure*}
\centering
\includegraphics[width=\textwidth]{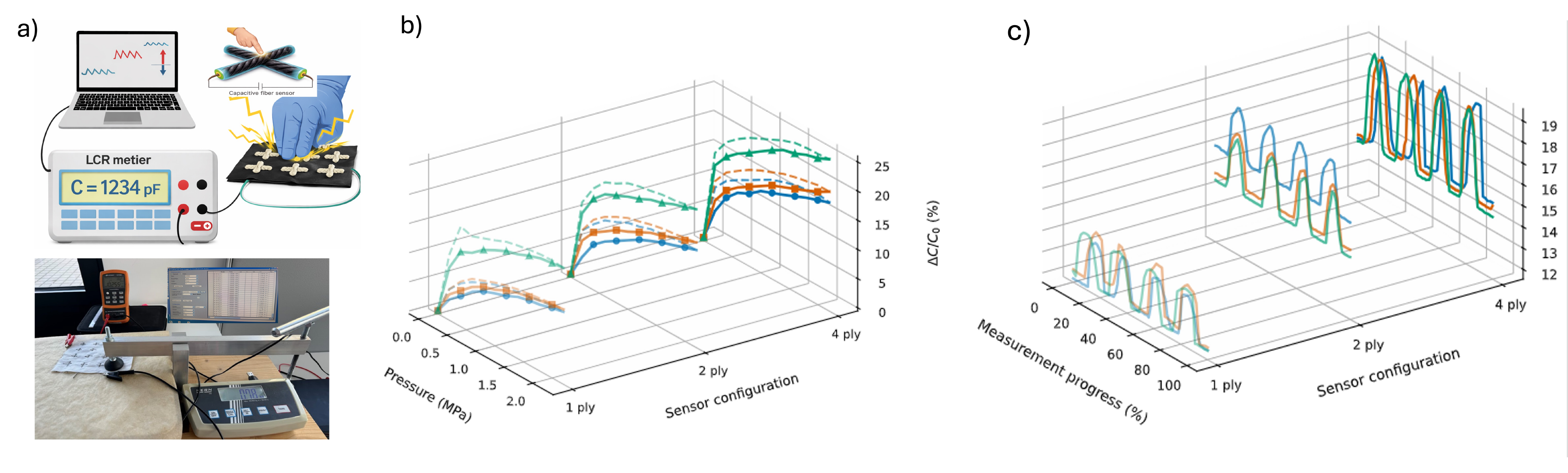}
\caption{Hysteresis and cyclic pressure response of the textile capacitive sensor. (a) Capacitive fiber sensing junction and LCR-meter measurement configuration. (b) Hysteresis across ply configurations. (c) Frequency response across ply configuration for dynamic loading and unloading.}
\label{figure_4}
\end{figure*}

Fig.~\ref{figure_4}(a) presents the sensing junction, experimental setup with LCR-meter connection and fabricated 4×4 textile arrays. Fig.~\ref{figure_4}(b) presents the 3D figure of quasi-static loading–unloading response obtained by applying and removing masses up to 1 kg. All configurations exhibit reversible capacitance changes with minimal baseline drift. This stable recovery is consistent with the elastic deformation and recovery mechanisms commonly reported for soft and textile-based capacitive pressure sensors \cite{Ha2022, su2022, Qu2023}.

Fig.~\ref{figure_4}(c) shows the pressure-dependent normalized capacitance response, $\Delta C / C_0$ of the one-, two-, and four-layer sensors during continuous dynamic loading–unloading at 1, 10, and 100 kHz. 
The nearly overlapping loading and unloading curves indicate low hysteresis and good mechanical recovery. The response increases with pressure and layer number, with the four-layer sensor producing the highest $\Delta C / C_0$. The slightly greater hysteresis observed in the multilayer structures may be associated with their more complex deformation and recovery behavior. Elasticity of the constituent fibers and dielectric layer is known to influence hysteresis in fiber-based capacitive pressure sensors \cite{su2022, Qu2023}.

The response amplitude increases from the one-layer to the four-layer configuration, confirming that the multilayer structure enhances the capacitive response while preserving repeatable operation.

\subsection{Environmental Stability and Proximity Sensing}

\subsubsection{Temperature Stability}

The thermal stability of the sensor was evaluated by monitoring the baseline capacitance while varying the ambient temperature from 25 to 90~$^\circ\mathrm{C}$. The experimental setup and thermal images are shown in Fig.~\ref{figure_5}(a–b), while the capacitance response is presented in Fig.~\ref{figure_5}(c). All sensor configurations exhibit only minor fluctuations in capacitance throughout, with no observable systematic drift or irreversible change. This stability is associated with the chemical inertness and low thermal expansion of the crosslinked PDMS dielectric layer \cite{Ha2022, su2022, Qu2023, Ye2022}. The results demonstrate stable sensor operation over the investigated temperature range without active compensation.

\subsubsection{Proximity Sensing}

\begin{figure*}
\centering
\includegraphics[width=\textwidth]{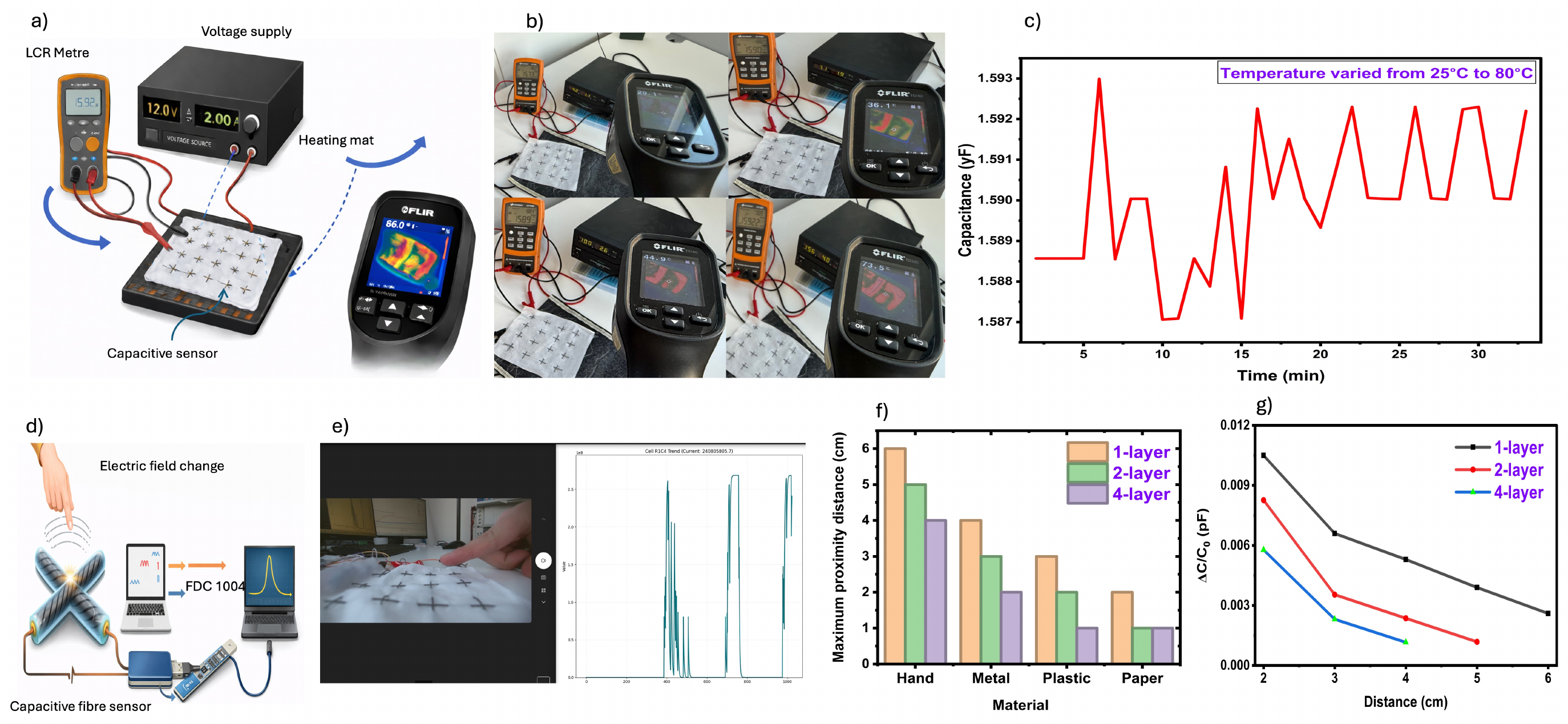}
\caption{Environmental stability and proximity sensing. (a-c) Capacitance measurements at different temperatures within 25–90~$^\circ\mathrm{C}$. (d-e) Fringe field perturbation schematic during proximity sensing. (f) Maximum proximity detection distance for different material types (paper, hand, metal, and plastic). (g) $\Delta C/C_0$ as a function of hand distance (0–6 cm) for all configurations.}
\label{figure_5}
\end{figure*}

The proximity-sensing capability was evaluated as a function of object distance and material type, as shown in Fig.~\ref{figure_5}(d–g). For all configurations, the capacitance increased as an object approached the sensor surface because of perturbation of the fringe electric field surrounding the conductive fiber electrodes \cite{Niu2024, Wu2025, Wu2024, Ye2022}.
As shown in Fig.~\ref{figure_5}(f), the human hand and metal plate produced larger capacitance changes than the plastic sheet and paper because conductive and higher-permittivity objects interact more strongly with the fringe electric field \cite{Niu2024, Wu2024, Wu2025}.
As shown in Fig.~\ref{figure_5}(g), the maximum detection distance decreased with increasing fiber number, from 60 mm for the 1-layer configuration to 50 mm for the 2-layer configuration and 40 mm for the 4-layer configuration. This trend may result from partial confinement of the fringe electric field by the more compact geometry and increased dielectric volume of the higher-layer yarns.

These results demonstrate dual-mode sensing within a single textile structure: contact-pressure detection and non-contact proximity detection. Such precontact sensing is particularly useful for robotic skin and human–robot collaboration, where early object detection can support collision avoidance and safer interaction \cite{Ham2022, Zahir2026}.

\subsection{Application Demonstrations on Robotic Platforms}
\subsubsection{System-Level Evaluation }
\begin{figure*}
\centering
\includegraphics[width=\textwidth]{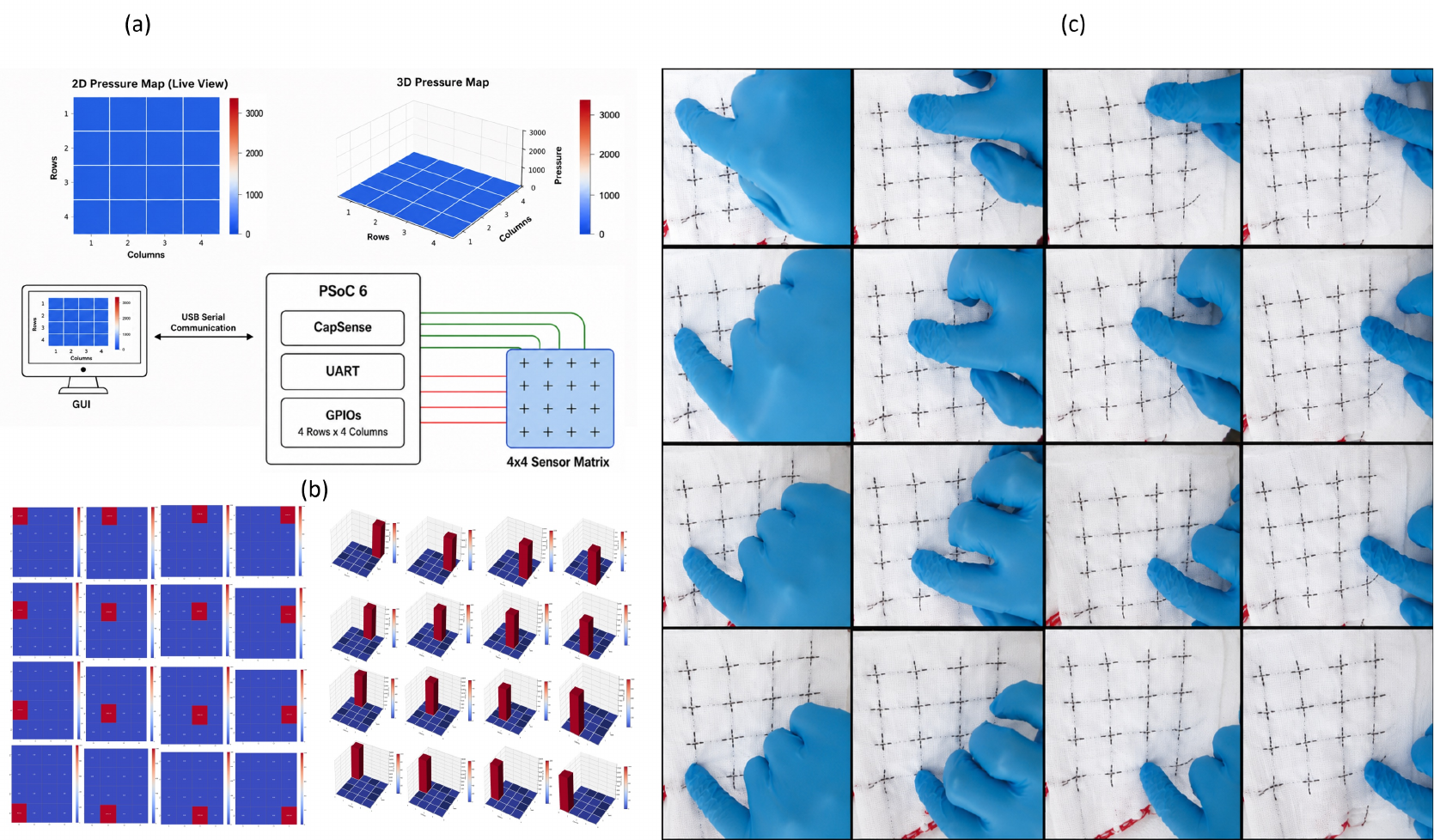}
\caption{Overview and validation of the $4\times4$ pressure-sensing matrix. 
(a) System architecture. 
(b) 2D/3D pressure maps for different touch positions. 
(c) Real-time finger-press pressure localization.}
\label{fig:pressure_matrix}
\end{figure*}

Fig.~\ref{fig:pressure_matrix} shows the $4 \times 4$  textile matrix used for localized touch and proximity detection. Each intersection of the dielectric-coated row and column yarns forms a sensing node. The combined row and column responses were used to identify the interaction location.
The measured signals were displayed as 2D heatmaps and 3D surface plots, representing the spatial position and intensity of the detected event. The results confirm the matrix’s ability to provide real-time distributed tactile and proximity sensing for human–robot interaction. The real-time response of the matrix to localized touch is presented in Supplementary Video~S1.

\subsubsection{Evaluation and Integration on a Robotic Arm}
\begin{figure*}
\includegraphics[width=\textwidth]{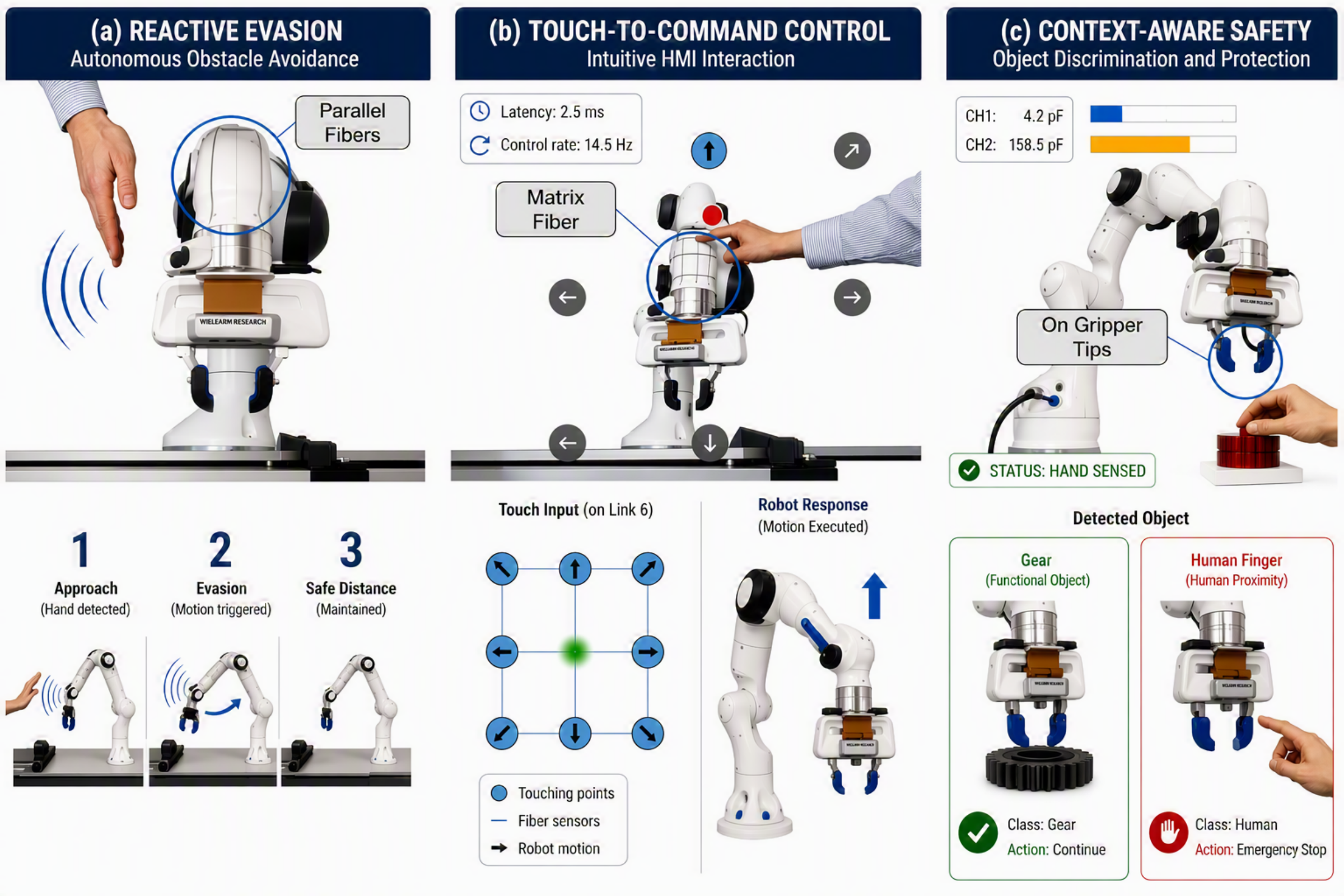}
\caption{Experimental validation of the textile sensing platform: 
(a) Autonomous evasion triggered by proximity detection; 
(b) Interactive touch localization for intuitive robot control; 
(c) Intelligent classification distinguishing between metallic gears and human finger proximity.}
\label{fig:demonstrations}
\end{figure*}

To evaluate its viability as an artificial e-skin, we implemented three distinct demonstrations, illustrating its potential for reflexive reactive motion, intuitive human machine interface (HMI), and context-aware safety monitoring. 
Fig.~\ref{fig:demonstrations}(a) illustrates the sensor's capacity for real-time reactive motion. 
By leveraging the fringe electric field, the robotic arm detects an approaching human hand and executes an autonomous ``evasion'' maneuver, demonstrating reflexive behavior relevant to safe operation in shared workspaces. The complete demonstration is provided in Supplementary Video~S2.
As shown in Fig.~\ref{fig:demonstrations}(b), the sensor functions as an intuitive Human-Machine Interface (HMI), where localized touch inputs are translated into directional commands for a robotic manipulator. The complete HMI demonstration is provided in Supplementary Video~S3.
Finally, Fig.~\ref{fig:demonstrations}(c) highlights the system's intelligent safety capabilities. Here capacitance thresholds are used for object classification. The threshold-based classification rule distinguishes between a gear and a human hand. While the gear engagement is identified as a functional task, human finger proximity is classified as an anomaly, triggering an emergency halt. The end-to-end robotic reaction latency, including sensing, signal processing, communication, and robot actuation, was approximately 403 ms. These demonstrations show that a single sensing platform, when coupled with appropriate software logic, can address multiple requirements of collaborative industrial robotics. This safety-monitoring demonstration is presented in Supplementary Video~S4.

\section{Conclusion}
This work presents a textile-integrated capacitive sensing platform in which twisted-yarn layer architecture serves as the primary structural design parameter governing sensing and measurement characteristics. 
By assembling PDMS-coated silver-coated conductive yarns into 1-layer, 2-layer, and 4-layer twisted configurations, we demonstrate that the effective electrode overlap area, $A_{\mathrm{eff}}$, at the fiber–fiber contact junction scales systematically from 0.25 to 1.0 to 4.0 mm², corresponding to a 16-fold increase from the 1-layer to the 4-layer configuration. The results demonstrate a quantitative relationship between yarn architecture, effective electrode overlap area, and the measured capacitive response. 
This direct geometric quantification of $A_{\mathrm{eff}}$ as a function of yarn architecture, validated by cross-sectional optical microscopy, 
provides a quantitative relationship between yarn architecture, effective electrode overlap area, and capacitive sensing characteristics in fiber-based capacitive sensors.

The optimized 4-layer configuration delivers the highest pressure sensitivity across applied stresses of 0.4–3.9~$\mathrm{MPa}$, stable cyclic operation over 15,000 loading cycles, low hysteresis, a rise and recovery time of approximately 0.9~s. Although the rise and recovery times under step loading are approximately 0.9 s, periodic excitation experiments demonstrated detectable/reproducible modulation rather than full settling at excitation frequencies up to 5 Hz. Thermal characterization confirms minimal capacitance drift across 25–90~$^\circ\mathrm{C}$ , a direct consequence of the chemically inert and elastomerically stable PDMS dielectric. Proximity sensing experiments reveal a layer-dependent detection range of 60, 50, and 40 mm for the 1-layer, 2-layer, and 4-layer configurations, respectively, exposing a fundamental sensitivity–range trade-off intrinsic to the yarn geometry: increasing layer number enhances contact pressure sensitivity through greater electrode overlap while simultaneously confining the fringe electric field and reducing pre-contact detection range. System-level demonstrations on a Franka Emika Panda robotic arm confirm reliable spatial mapping of single- and multi-point contact events and real-time dual-mode proximity and touch detection for human–robot interaction.

Collectively, these results establish yarn layer architecture as a scalable, tunable, and fabrication-compatible design parameter for multifunctional textile-based capacitive sensors, offering architecture-dependent tuning of contact sensitivity, proximity range, mechanical strength, and thermal stability through a single structural variable. 
These findings provide guidance for the design and optimization of textile-integrated capacitive sensing systems for wearable electronics, electronic skin, and robotic applications.

\section*{ACKNOWLEDGMENT}
The authors acknowledge the financial support of the SmartSense AI Eurostars project (Project No. 3087), co-funded through the Eurostars programme by the European Commission and the national funding agencies of the participating countries. The Austrian project partners were supported by the Austrian Federal Ministry for Innovation, Mobility and Infrastructure (BMIMI) and managed by the Austrian Research Promotion Agency (FFG) (Project ID: FFG 904634), while the Swiss project activities at ZHAW Zurich University of Applied Sciences were supported by Innosuisse, the Swiss Innovation Agency. The project was further supported through partner co-investment and in-kind contributions from the participating organizations. Generative artificial intelligence (AI) tools were used solely for language refinement and editorial assistance during manuscript preparation.


\ifCLASSOPTIONcaptionsoff
  \newpage
\fi

\bibliographystyle{IEEEtran}
\bibliography{reference}

\newpage
\vspace{4pt}
\begin{IEEEbiography}[{\includegraphics[width=1in,height=1.25in,clip,keepaspectratio]{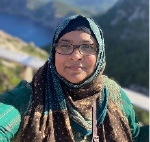}}]{Ishtia Zahir } received her Master’s in micro and 
nanoelectronics from University of Autonoma Barcelona and is 
currently working as research engineer at V-Trion GmbH in 
Austria.
\end{IEEEbiography}

\vspace{4pt}

\begin{IEEEbiography}[{\includegraphics[width=1in,height=1.25in,clip,keepaspectratio]{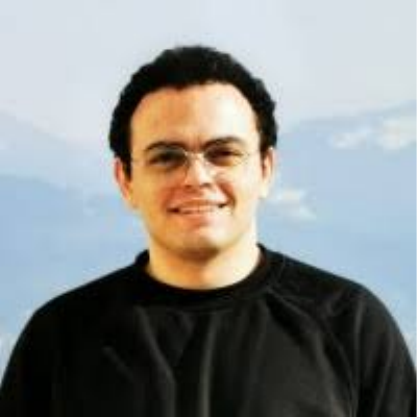}}]{Eslam Saleh } 
received the B.Sc. degree in Mechatronics Engineering from the German University in Cairo, Egypt, in 2020, and the M.Sc. degree in System Design, with a specialization in Mechatronics, from Carinthia University of Applied Sciences (FH Kärnten), Austria, in 2023. He is currently an Embedded Electronics and Software Engineer with V-Trion GmbH, Austria. His research interests include embedded systems, smart textile sensors, wearable electronics, biomedical sensing, and sensor-data acquisition and analysis.
\end{IEEEbiography}

\vspace{4pt}

\begin{IEEEbiography}
[{\includegraphics[width=1in,height=1.25in,clip,keepaspectratio]{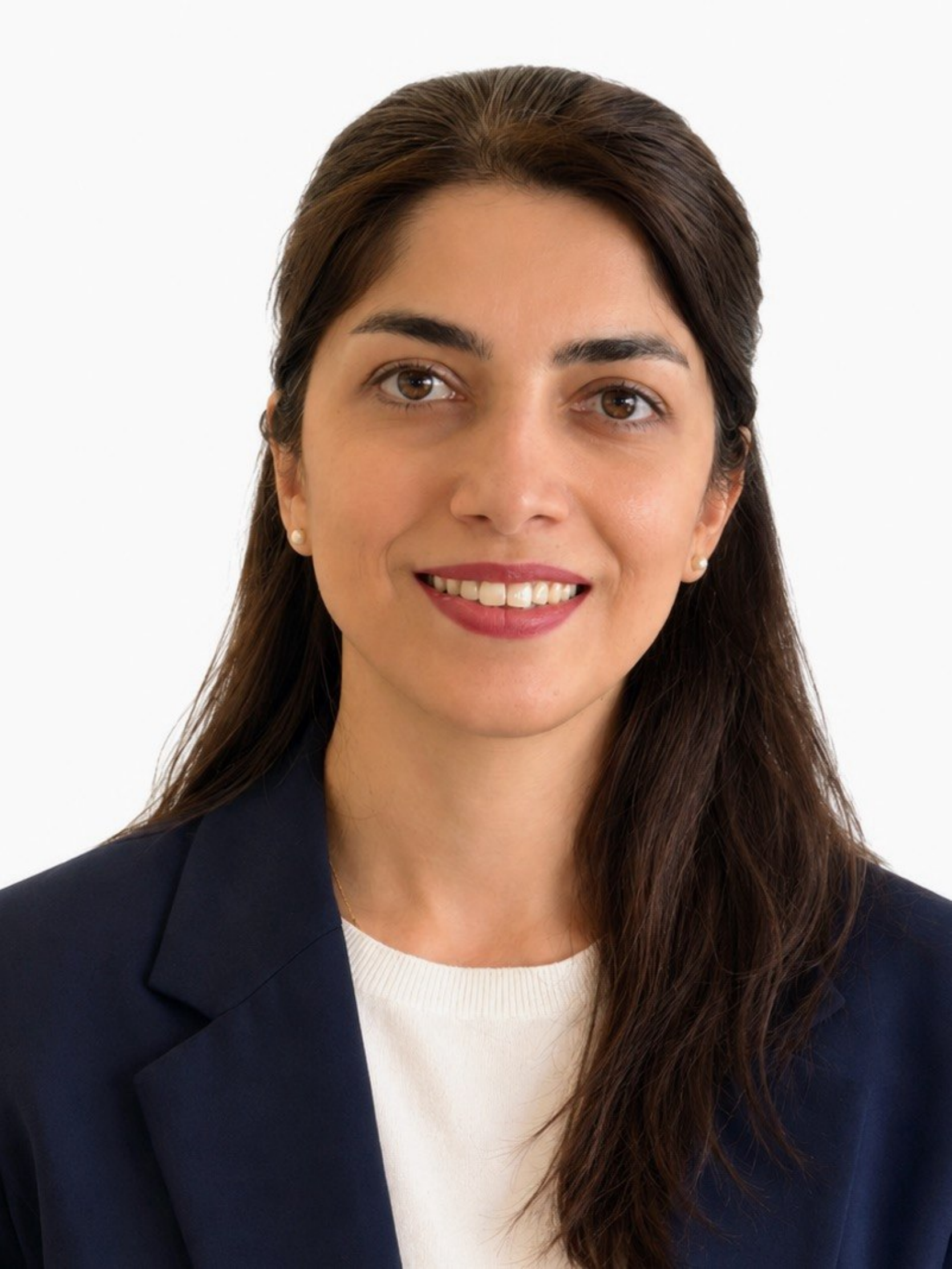}}]{Maryam Rezayati }
received the B.S. degree in robotics engineering from Hamedan University of Technology, Iran, in 2014, and the M.S. degree in mechatronics engineering from the University of Isfahan, Iran, in 2017. She defended her Ph.D. degree in Informatics at the University of Zurich, Switzerland, in 2025.
Since 2023, she has been a project manager and research associate with the Institute of Mechatronic Systems, Zurich University of Applied Sciences (ZHAW), Winterthur, Switzerland.
Her research interests include physical human-–robot interaction, learning--based robot perception, and AI-based industrial robotic applications.
\end{IEEEbiography}

\vspace{4pt}

\begin{IEEEbiography}[{\includegraphics[width=1in,height=1.25in,clip,keepaspectratio]{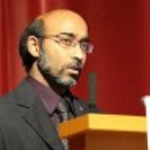}}]{Gaffar Hossain } received the European doctorate degree in 
polymer and biopolymer from the Technical University 
of Catalonia, Barcelona, Spain, in 2011, and the M.Sc. degree 
from Dresden University of Technology, Germany, in 2005. Since 2011, he has been the Head of V-Trion 
GmbH, Austria.
\end{IEEEbiography}

\vspace{4pt}

\begin{IEEEbiography}[{\includegraphics[width=1in,height=1.25in,clip,keepaspectratio]{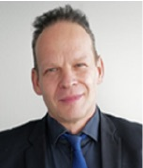}}]{ Günter Grabher } completed his diploma in textile technology 
from Höhere Technische Lehranstalt (HTL), Dornbirn 
(Austria) in 1990. He gained over 30 years of his experience in 
the field of textile technology as well as in smart textile. He is 
currently the founder of Grabher Group.
\end{IEEEbiography}

\end{document}